\documentclass[a4paper]{cas-sc}

\usepackage[numbers]{natbib}
\usepackage{graphicx}
\usepackage{multirow}
\usepackage[normalem]{ulem}
\usepackage{bbding}

\def\tsc#1{\csdef{#1}{\textsc{\lowercase{#1}}\xspace}}
\tsc{WGM}
\tsc{QE}

\begin{document}
\let\WriteBookmarks\relax
\def\floatpagepagefraction{1}
\def\textpagefraction{.001}
\let\printorcid\relax

\shorttitle{SA$^{2}$Net: Scale-Adaptive Structure-Affinity Transformation for Spine Segmentation from Ultrasound Volume Projection Imaging}    

\shortauthors{Hao Xie, et al.}  

\title [mode = title]{\texorpdfstring{SA$^{2}$Net: Scale-Adaptive Structure-Affinity Transformation for Spine Segmentation from Ultrasound Volume Projection Imaging}{SA2Net: Scale-Adaptive Structure-Affinity Transformation for Spine Segmentation from Ultrasound Volume Projection Imaging}}  



%

\author[1]{Hao Xie}

\cormark[1]


\ead{carry-h.xie@connect.polyu.hk}


\credit{Writing – original draft, Writing – review and editing, Investigation, Methodology, Validation, Visualization}

\affiliation[1]{organization={Department of Electrical and Electronic Engineering, The Hong Kong Polytechnic University},
            country={Hong Kong}}

\author[2]{Zixun Huang}
\ead{zixunhuang@szpu.edu.cn}
\credit{Writing – review and editing, Conceptualization, Formal analysis, Investigation}
\affiliation[2]{organization={School of Artificial Intelligence, Shenzhen Polytechnic University},
            city={Shenzhen},
            country={China}}

\author[1]{Yushen Zuo}
\ead{yushen.zuo@polyu.edu.hk}
\credit{Writing – review and editing, Methodology, Validation, Visualization}

\author[3]{Yakun Ju}
\ead{kelvin.yakun.ju@gmail.com}
\credit{Writing – review and editing, Validation, Visualization}
\affiliation[3]{organization={School of Computing and Mathematic Sciences, University of Leicester},
            country={the UK}}

\author[1]{Frank H. F. Leung}
\ead{frank-h-f.leung@polyu.edu.hk}
\credit{Supervision, Funding acquisition, Project administration, Resources, Conceptualization, Formal analysis}

\author[1]{N. F. Law}
\ead{ngai.fong.law@polyu.edu.hk}
\credit{Writing – review and editing, Validation, Supervision}

\author[1]{Kin-Man Lam}
\ead{kin.man.lam@polyu.edu.hk}
\credit{Writing – review and editing, Supervision, Resources}

\author[4]{Yong-Ping Zheng}
\ead{yongping.zheng@polyu.edu.hk}
\credit{Writing – review and editing, Supervision, Conceptualization, Data curation, Project administration}
\affiliation[4]{organization={Department of Biomedical Engineering, The Hong Kong Polytechnic University},
            country={Hong Kong}}

\author[5]{Sai Ho Ling}
\ead{steve.ling@uts.edu.au}
\credit{Writing – review and editing, Supervision, Conceptualization, Investigation}
\affiliation[5]{organization={School of Electrical and Data Engineering, University of Technology Sydney},
            country={Australia}}

\cortext[1]{Corresponding author}



\begin{abstract}
Spine segmentation, based on ultrasound volume projection imaging (VPI), plays a vital role for intelligent scoliosis diagnosis in clinical applications. However, this task faces several significant challenges. Firstly, the global contextual knowledge of spines may not be well-learned if we neglect the high spatial correlation of different bone features. Secondly, the spine bones contain rich structural knowledge regarding their shapes and positions, which deserves to be encoded into the segmentation process. To address these challenges, we propose a novel scale-adaptive structure-aware network (SA$^{2}$Net) for effective spine segmentation. First, we propose a scale-adaptive complementary strategy to learn the cross-dimensional long-distance correlation features for spinal images. Second, motivated by the consistency between multi-head self-attention in Transformers and semantic level affinity, we propose structure-affinity transformation to transform semantic features with class-specific affinity and combine it with a Transformer decoder for structure-aware reasoning. In addition, we adopt a feature mixing loss aggregation method to enhance model training. This method improves the robustness and accuracy of the segmentation process. The experimental results demonstrate that our SA$^{2}$Net achieves superior segmentation performance compared to other state-of-the-art methods. Moreover, the adaptability of SA$^{2}$Net to various backbones enhances its potential as a promising tool for advanced scoliosis diagnosis using intelligent spinal image analysis. The code and experimental demo are available at \textit{https://github.com/taetiseo09/SA2Net}.
\end{abstract}


\maketitle

\section{Introduction}
Scoliosis is a serious deformity of the spinal cord characterized by a spine curvature angle exceeding 10° 
\cite{Baner2024_1}. Adolescent idiopathic scoliosis (AIS) is the most prevalent form, affecting 3–4\% of children in Hong Kong \cite{Scolioscan_3}. AIS is typically diagnosed during the crucial adolescent growth period between 10 and 14 years of age \cite{Scolioscan_7}. If left untreated, AIS can cause permanent damage to the skeletal structure \cite{Baner2024_8}. In clinical practice, the Cobb Angle on radiographs \cite{Baner2024_11} is the most widely used measurement to quantify the magnitude of spinal deformities and is considered the gold standard for scoliosis diagnosis. However, radiation exposure poses significant health risks, particularly in childhood, where it has been linked to an increased risk of cancer and leukemia \cite{Scolioscan_14}. Consequently, there is an urgent need for a radiation-free imaging approach for assessing AIS.

Ultrasound, as a radiation-free and cost-effective imaging modality, holds significant potential for widespread use in scoliosis diagnosis. Recent studies have clearly demonstrated the feasibility of using three-dimensional (3D) ultrasound imaging methods to measure scoliotic deformity in vivo \cite{Scolioscan_31, Scolioscan_35, Scolioscan_34}. Volume projection imaging (VPI) \cite{Scolioscan_32} was proposed to generate two-dimensional (2D) coronal view images of spine structure by projecting voxels from 3D ultrasound volume data onto a 2D plane. This process involves averaging the intensity of all voxels to create an X-ray-like image in the coronal plane. This enables precise localization of critical bony features, making the assessment of spinal deformities possible on this 2D plane. The Ultrasound Curve Angle (UCA) \cite{Zixun_joint6}, analogous to the radiographic Cobb Angle \cite{Baner2024_11}, is applied for evaluating spine curvature. Accurate calculation of the spinal curve angle requires the segmentation of ribs, thoracic processes, and the lump (Figure \ref{fig1}(c)). Spine segmentation from ultrasound VPI images serves as a crucial pre-processing step for the automatic measurement of spinal deformities, providing the basis for intelligent scoliosis diagnosis.

Over the past few years, rapid advancements in artificial intelligence (AI) and deep learning have opened new avenues for automatic medical image segmentation based on extensive experimental research. Existing medical image segmentation methods can primarily be divided into two categories: those based on convolutional neural networks (CNNs) and those based on Transformer networks. Inspired by the encoder-decoder architecture and fully convolutional network (FCN) \cite{FCN}, a variety of U-shaped CNN architectures, such as UNet \cite{UNet} and UNet++ \cite{UNet++}, have become standard for high-quality segmentation. The success of CNNs is largely attributed to the scale invariance and inductive bias of the convolution operation. However, CNNs are limited in their ability to capture the relationship between distant pixels in medical images.

To address this shortcoming, researchers have proposed a Transformer architecture \cite{transformer} based on the self-attention mechanism. This architecture can process indefinite-length input, establish long-range dependencies, and capture global information. The hybrid architecture, combining CNN and Transformer, leverages the strengths of both approaches to model local details and global semantic information in medical images, thus achieving better segmentation results \cite{Azad_2022}. Nevertheless, existing segmentation methods still face challenges that hinder their performance in spinal ultrasound VPI image segmentation.

First, the correlation between the feature map space and the channels is often overlooked \cite{Hong_2021}, leading to inadequate global semantic feature expression of spine contextual knowledge. Additionally, spine bones follow a relatively uniform position and shape, appearing only in specific regions of one ultrasound image. The strong prior knowledge of these shapes and positions deserves thorough analysis. More importantly, due to insufficient consideration of the high spatial correlation between different bone features, adjacent tissue parts are blended in the input space and may obfuscate the segmentation model, leaving ambiguous segment boundaries. One potential solution is reducing reliance on pixel-level semantic information and incorporating additional structural knowledge to separate similar and entangled representations. Therefore, we empirically hypothesize that enforcing a strict constraint on spine structural information learning can enhance ultrasound VPI image processing.

In this paper, we propose a novel scale-adaptive structure-aware network, termed SA$^{2}$Net, for the enhancement of spine segmentation in ultrasound VPI images. Firstly, we design a scale-adaptive channel-spatial attention module (SACSAM) to achieve cross-dimensional global modeling of spinal images. SACSAM consists of two parallel branches, extending the dual attention block \cite{SCSA-Net}. The learnable scale parameters are applied to fully compensate for the limitations of the conventional dual attention mechanism in modeling the cross-dimensional relationship between channel and spatial dimensions. Through these two parallel branches of scale-adaptive learning, SACSAM captures richer long-distance correlation features and enhances the extraction of long-range spine contextual representations. Secondly, we propose Structure-Affinity Transformation to transform semantic features with class-specific affinity that encode structural information. This transformation brings features from the same category closer together while pushing features from different categories apart. We find that the multi-head self-attention in Transformers can capture semantic-level affinity \cite{affinity} and can be used to learn structural knowledge of different spine bones. Thus, based on the proposed structure-affinity transformation, we employ a Transformer module \cite{transformer}, called structure-aware module (SAM), to impose structure-affinity attention weights. This process highlights relevant feature maps and facilitates structure-aware reasoning. Consequently, our proposed SA$^{2}$Net effectively fuses cross-dimensional features and class-specific structure-affinity features, generating the final prediction in the decoder.

In addition, to effectively encode contextual and structural knowledge of the spine in SA$^{2}$Net, inspired by the work of MERIT \cite{MERIT}, we adopt a feature mixing loss aggregation method for spine segmentation learning. This method enables better model training by automatically supervising the feature maps extracted from SACSAM and SAM, calculating the loss across all prediction combinations to optimize segmentation from detailed features to the overall structure.
To summarize, our contributions are as follows:

\begin{itemize}
\item We introduce a scale-adaptive strategy for the dual spatial-channel attention block, enabling two parallel branches to complement each other and capture cross-dimensional long-range dependencies between channel and spatial dimensions.

\item We propose Structure-Affinity Transformation and wrap it into a structure-aware module to enhance structure-affinity feature representation ability and improve the separability between the segmented bone features.

\item We apply a feature mixing loss aggregation approach to generate new synthetic predictions and then propose a novel scale-adaptive structure-aware network (SA$^{2}$Net) for spine segmentation from ultrasound VPI images.

\item We conduct rigorous experiments and ablation studies on scoliosis data to show that SA$^{2}$Net is compatible with both CNN and Transformer backbones and produces new state-of-the-art results with the Swin-Transformer \cite{swin} backbone, revealing the superiority of our segmentation network.
\end{itemize}

The remainder of this paper is organized as follows: Section \ref{related} summarizes related work about our proposed methods. Section \ref{method} describes the proposed SA$^{2}$Net architecture and methodology. Section \ref{experiment} gives an account of the experimental setups and the analysis of the experimental results. Next, Section \ref{discussion} discusses the clinical value of this research and possible limitations faced in the practical application. Finally, Section \ref{conclusion} concludes this paper.

\section{Related Work} \label{related}
\subsection{Ultrasound Volume Projection Imaging}
\begin{figure}[t]
\centerline{\includegraphics[keepaspectratio, width=0.45 \linewidth]{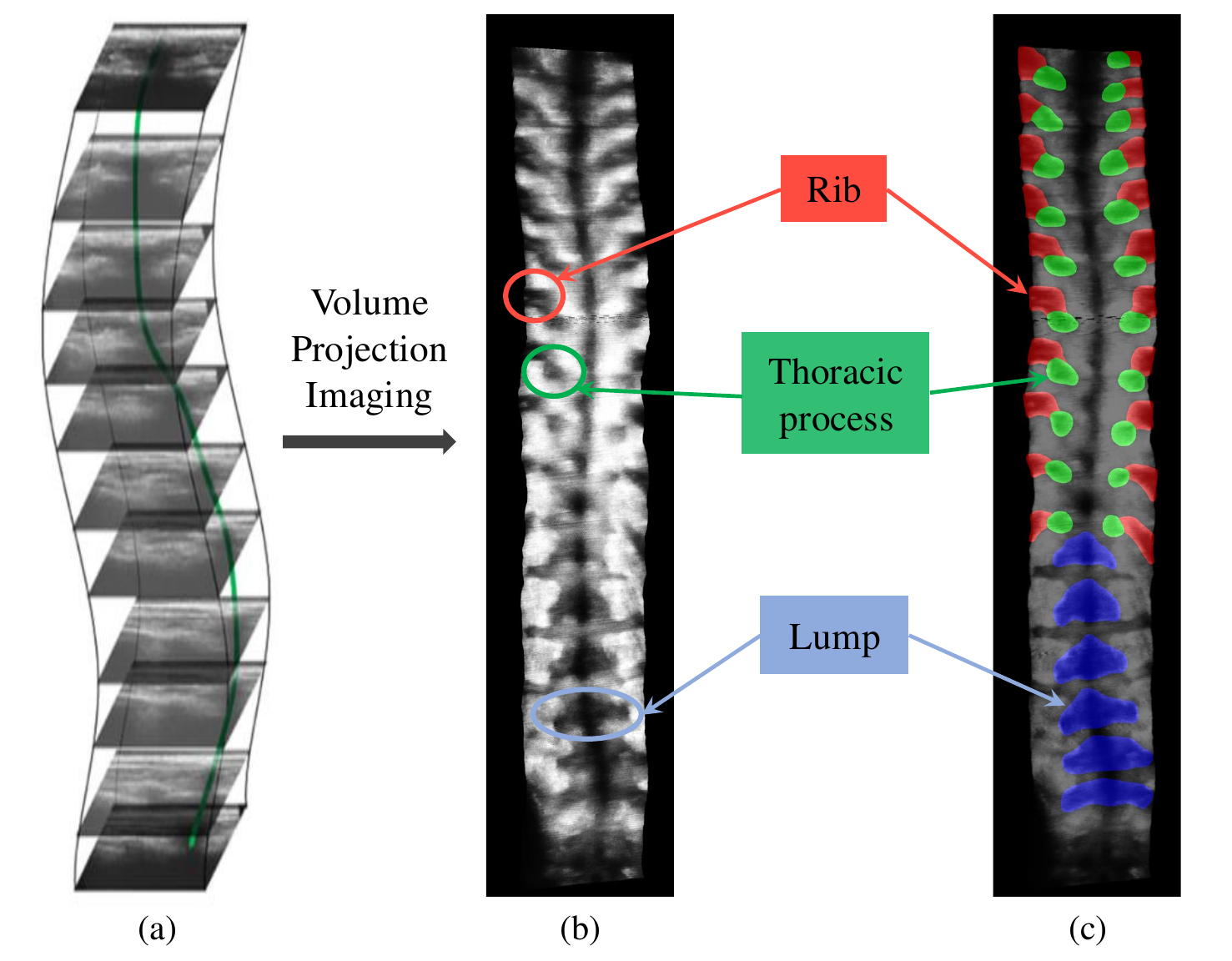}}
\caption{An illustration of spine segmentation from ultrasound VPI images. (a) Spinal 3D ultrasound volume data, which are in the form of an ultrasound sequence of 2D slices; (b) One extracted ultrasound VPI image based on the projection on the 2D coronal plane; and (c) Different bone features in the spinal image. The segmented rib and thoracic process are painted red and green, respectively. The lump, which is formed by the combined shadow of the partial bilateral inferior articular process, laminae, and the superior articular process of the inferior vertebrae, is painted blue.}
\vspace{-5mm}
\label{fig1}
\end{figure}

The feasibility and reliability of AIS diagnosis using ultrasound imaging have been well established \cite{Scolioscan_34, Scolioscan_35}. Volume projection imaging is a practical visualization technique that leverages 3D ultrasound volume data to display spinal deformities on the 2D coronal plane \cite{Scolioscan_32}. The working pipeline is shown in Figure \ref{fig1}. The assessment procedure involves freehand scanning with the probe along the patient’s back, covering the entire spine area. After scanning, the acquired B-mode image data, along with the corresponding position and orientation information, are used to reconstruct a 3D image of the coronal view of the spine. The core of VPI is to average the intensity of all voxels within a selected depth along the anterior-posterior axis to form a 2D image on the coronal plane. The spine curvature angle, named Scolioscan Angle \cite{Scolioscan}, is then derived according to the orientations of the two lines drawn on the image. Ultrasound curve angle measurement requires precise delineation of thoracic and lumbar bony features from the ultrasound VPI image. That means accurate medical image segmentation has become a key component of computer-aided scoliosis diagnosis.

\subsection{Approaches of Spine Segmentation with Ultrasound}
In medical image segmentation, UNet and its family, such as UNet++ \cite{UNet++}, ResUNet \cite{ResUNet}, and nnUNet \cite{nnUNet}, are widely used due to their simple but effective encoder-decoder design. Ungi \emph{et al.} \cite{Zixun_joint21} were the first to propose an end-to-end framework for spine segmentation of 2D ultrasound images. However, as this method was applied to sparse 2D images, the predicted segments were of low accuracy. Thanks to volume projection imaging, this 3D volume compression technique, more reliable approaches were proposed to visualize the entire spine anatomy. Huang \emph{et al.} \cite{RSN-UNet} improved segmentation robustness to speckle and regular occlusion noise in VPI images by introducing a total variance loss into UNet. Further advancement came with SIU-Net \cite{SIU-Net}, which incorporated an improvised inception block and newly designed decoder-side dense skip pathways.

Recent studies have explored the integration of attention mechanisms within the encoder or decoder for enhancing spine segmentation from ultrasound VPI images. Zhao \emph{et al.} \cite{Zixun_joint24} introduced structure supervision to the representation learning in a self-attention manner. A dual-task framework \cite{Zixun_joint25} with boundary detection as an auxiliary task was presented to regularize spine segmentation. Huang \emph{et al.} \cite{Zixun_joint} proposed a novel joint learning network capable of simultaneously performing noise removal and spine segmentation. In our previous study, we have proposed a novel structure-affinity dual attention-based network (SADANet) \cite{BIBM} to segment spinal ultrasound images. Despite these advancements, all the above methods still employed UNet or ResNet \cite{Zixun_joint46} as the backbone. 

With the emergence of Vision Transformers (ViT) \cite{ViT}, segmentation models based on Transformer and CNN-Transformer, such as TransUNet \cite{TransUNet} and Swin-UNet \cite{Swin-UNet}, have begun to be applied to medical image segmentation. To the best of our knowledge, although Transformer-based architectures showed promising results in spinal X-ray or computed tomography (CT) image segmentation, very few explorations have been made on spine segmentation from ultrasound VPI images \cite{ISBI}. This gap motivates us to investigate a more versatile network in this paper that can perform spine bone segmentation.

\section{Methodology} \label{method}
\begin{figure*}[t]
\centering
\includegraphics[width=0.95 \linewidth]{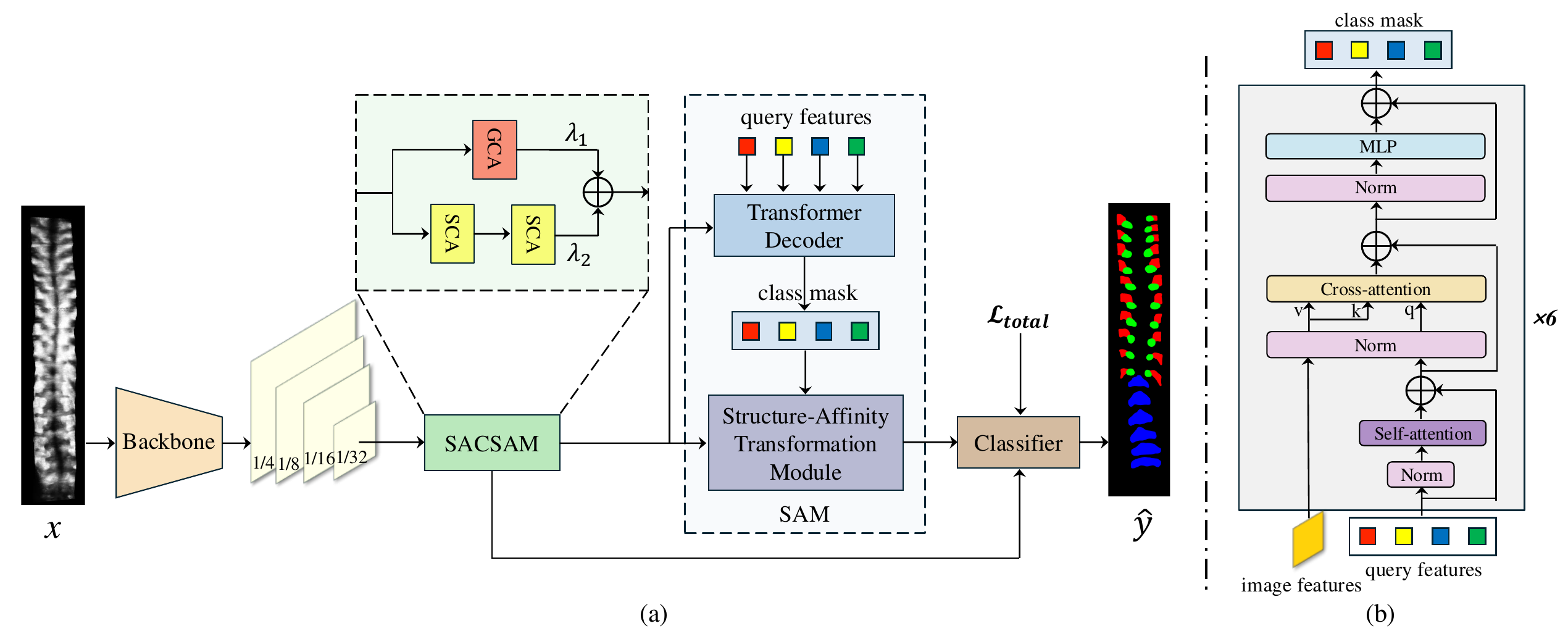}
\caption{(a) An overview of the proposed SA$^{2}$Net. $x$ represents an input spinal image while $\hat{y}$ denotes the predicted segmentation result. $\mathcal{L} _{total}$ is the optimized loss during the training process; (b) An illustration of the specially designed Transformer decoder with cross-attention. The query features are input into this Transformer decoder and are updated with multi-scale image features, generating semantic class masks.}
\vspace{-5mm}
\label{SA2Net}
\end{figure*}

In this section, we describe our proposed scale-adaptive structure-aware network, named SA$^{2}$Net, for spine segmentation from ultrasound VPI images. We start with a comprehensive overview of the framework, illustrated in Figure \ref{SA2Net}(a). Next, we provide details for key components of this architecture, including the scale-adaptive strategy for channel and spatial attentions, the structure-affinity transformation mechanism, and how the structure-aware module leverages this mechanism. Finally, we introduce the loss function design based on a feature mixing loss aggregation approach.

\subsection{Overview of SA$^{2}$Net}
Figure \ref{SA2Net}(a) presents the overall pipeline of SA$^{2}$Net, which is built upon an end-to-end segmentation architecture. SA$^{2}$Net consists of a backbone, a scale-adaptive channel-spatial attention module, a structure-aware module, and multiple prediction heads to produce the segmentation outputs. Given a spinal ultrasound VPI image, we extract multi-scale features with backbones like ResNet \cite{Zixun_joint46} or Swin-Transformer \cite{swin}. The backbone choice of SA$^{2}$Net is flexible, and our proposed modules can be easily integrated into any encoder-decoder architecture. Subsequently, we adopt the scale-adaptive channel-spatial attention module (SACSAM), which comprises a global channel attention (GCA) module and two consecutive spatial criss-cross attention (SCA) modules in a parallel manner. The learnable scale parameters and an element-wise addition are applied to capture multi-scale spatial information for each feature channel, effectively integrating bone feature dependencies between the channel and spatial dimensions and performing multi-scale feature fusion.

More importantly, after feature enhancement with SACSAM, the structure-aware module (SAM) combines our proposed structure-affinity transformation with a specially designed Transformer decoder (Figure \ref{SA2Net}(b)) that exploits structural information in semantic features. The Transformer decoder utilizes the cross-attention mechanism to combine the image features from SACSAM and update the queries layer-by-layer for better structure reasoning. The structure-affinity transformation module then transforms processed semantic features with class-specific affinity by encoding structural knowledge of different bone regions into structure-affinity attention weights thereby enhancing spine structural distinctions across segment categories. Finally, the feature maps from two parts of SA$^{2}$Net are fed to a classifier together for classification, and the spine bone feature classification results are obtained using a combinatorial learning strategy to optimize the spine segmentation model during training.

\begin{figure*}[t]
\centering
\includegraphics[width=0.95 \linewidth]{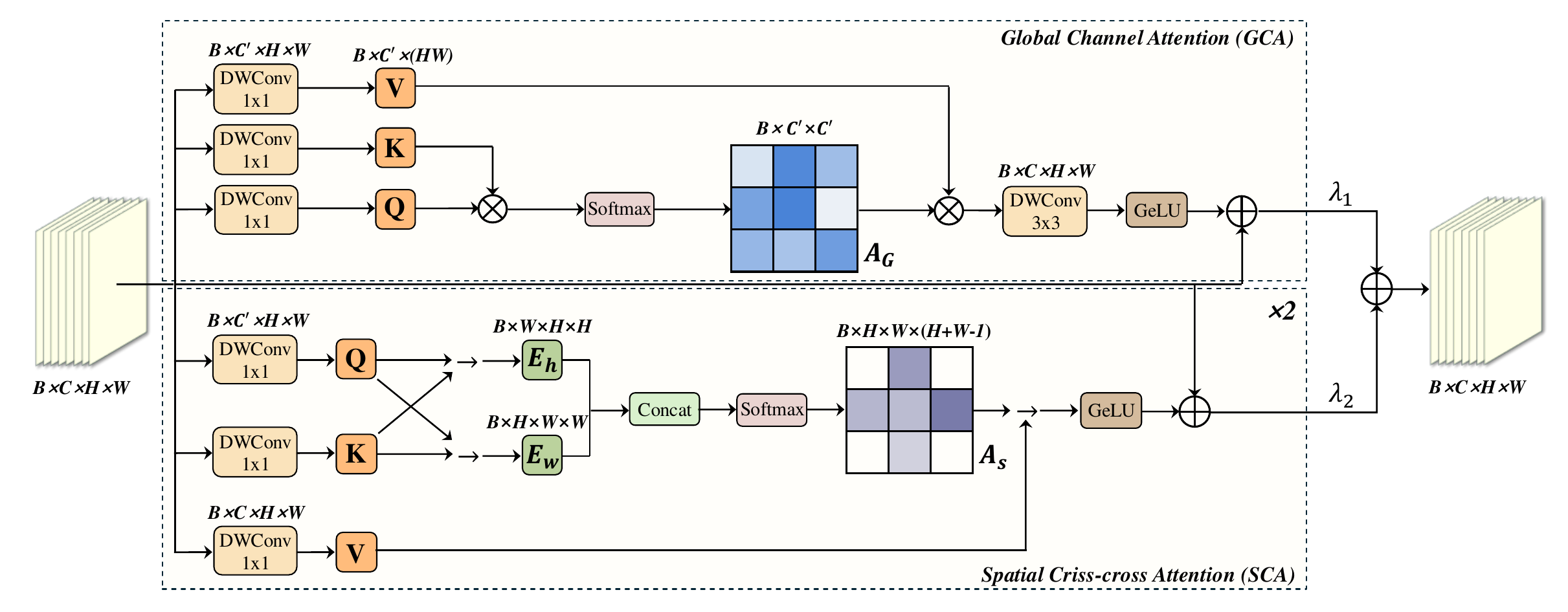}
\caption{Details of the scale-adaptive channel-spatial attention module (SACSAM). $B$ denotes the batch size, $C$ represents the number of channels, and $H$ and $W$ correspond to the height and width of the input feature map $X$, respectively.}
\vspace{-5mm}
\label{SACSAM}
\end{figure*}

\subsection{Scale-Adaptive Channel-Spatial Attention}
The attention mechanism captures long-range dependencies of feature maps and enhances representations of interest, facilitating the learning of more discriminative features. However, it only considers the dependency in the spatial dimension and not the cross-dimensional dependency between spatial and channels \cite{Hong_2021}. Therefore, when dealing with spine segmentation from ultrasound VPI images, the attention mechanism ignores the correlation between spatial dependencies and channel relationships, resulting in poor global semantic feature expression. To solve these problems, we propose a novel scale-adaptive channel-spatial attention module called SACSAM. As depicted in Figure \ref{SACSAM}, the details of our module are described next.

\subsubsection{Global Channel Attention Module}
One of the two parallel branches is composed of a global channel attention (GCA) module \cite{BIBM}. Since channel maps can be treated as class-specific multi-spatial responses, and some relatively important channels usually have similar spatial responses, we employ the GCA module to assign different levels of importance to each channel, thus emphasizing more relevant features while suppressing less useful ones. Given the input $X \in \mathbb{R}^{B\times C\times H\times W}$, we reduce the number of channels to $C^{'}$, where $C^{'} = \frac{C}{8}$, using a 1$\times$1 depth-wise convolution to reduce computational complexity. Afterwards, we recover the original channels using another 3$\times$3 depth-wise convolution, followed by a GeLU activation. Inspired by the significant advantages of the ViT \cite{ViT} in utilizing multi-head self-attention for modeling similarities, we leverage this mechanism to generate the channel dependency matrix $A_{G} \in \mathbb{R}^{B\times C^{'}\times C^{'}}$, and then the GCA-enhanced features $X_{c}$, finally obtained by an element-wise addition operation with the input feature map $X$. The implementation process is defined as follows:
\begin{equation}
\begin{aligned}
F_{proj\_1} = DWConv _{(1,1)} ^{C\to C^{'}}, F_{proj\_3} = DWConv _{(3,3)} ^{C^{'}\to C}
\end{aligned}
\end{equation}
\begin{equation}
\begin{aligned}
Q = F_{proj\_1}^{Q} (X), K = F_{proj\_1}^{K} (X), V = F_{proj\_1}^{V} (X)
\end{aligned}
\end{equation}
\begin{equation}
\begin{aligned}
X_{attn} = Attention(Q,K,V) = Softmax(\frac{Q K^{T}}{\sqrt{C^{'}}})V
\end{aligned}
\end{equation}
\begin{equation}
\begin{aligned}
X_{c} = X \oplus \Phi(F_{proj\_3} (X_{attn}))
\end{aligned}
\end{equation}
where $F_{proj\_1}(\cdot)$ and $F_{proj\_3}(\cdot)$ represent the depth-wise convolutions with kernel sizes of 1 and 3, respectively, while $\Phi(\cdot)$ denotes the GeLU normalization. In the GCA module, self-attention is computed along the channel dimension, where $Q,K,V\in \mathbb{R}^{B\times C^{'}\times (HW)}$.

\subsubsection{Spatial Criss-cross Attention Module}
Spatial attention determines where to focus in a feature map. This process enhances the ability to recognize and respond to the high spatial correlation of spine bones. To model full-image contextual dependencies using lightweight computation, we utilize the spatial criss-cross attention (SCA) module \cite{BIBM} to collect contextual information in horizontal and vertical directions. For each position in the feature map, only sparse connections $(H + W -1)$ are considered by aggregating features only in horizontal and vertical directions. Thus, two consecutive SCA modules are stacked as the other parallel branch to harvest full-image spine contextual information. This architecture compensates for the deficiency of criss-cross attention \cite{CCNet}, which cannot obtain dense contextual information from all pixels, and achieves more accurate segmentation performance for the spine bone with slightly more computational complexity. In our SCA module, depth-wise convolutions are also utilized to generate the query, key, and value. To generate feature maps along the $H$ and $W$ dimensions, respectively, we refer to the calculation of attention weight in the self-attention mechanism and perform the Einstein summation (einsum) operations between queries and keys in our practice. Next, we concatenate and apply a Softmax layer on them to obtain the spatial attention map $A_{S} \in \mathbb{R}^{B\times H\times W\times (H + W -1)}$. The einsum operation is utilized again between the spatial attention map and the key feature to finally generate the SCA-enhanced feature $X_{s}$. The whole process for extracting spatial information is shown as follows:
\begin{equation}
\begin{aligned}
F_{proj\_1} = DWConv _{(1,1)} ^{C\to C^{'}}, F_{proj\_3} = DWConv _{(3,3)} ^{C^{'}\to C}
\end{aligned}
\end{equation}
\begin{equation}
\begin{aligned}
Q = F_{proj\_1}^{Q} (X), K = F_{proj\_1}^{K} (X), V = DWConv _{(1,1)} ^{C\to C} (X)
\end{aligned}
\end{equation}
\begin{equation}
\begin{aligned}
E_{h} = (Q \to K)^{H}, E_{w} = (Q \to K)^{W}
\end{aligned}
\end{equation}
\begin{equation}
\begin{aligned}
A_{S} = Softmax(Concat(E_{h},E_{w}))
\end{aligned}
\end{equation}
\begin{equation}
\begin{aligned}
X_{s} = X \oplus \Phi(V\to A_{S})
\end{aligned}
\end{equation}
where the einsum operations are denoted as ``$\to$'', $E_{h} \in \mathbb{R}^{B\times W\times H\times H}$ and $E_{w} \in \mathbb{R}^{B\times H\times W\times W}$. It is vital to note that dimension reduction is only applied to the generation of query and key features, while $V\in \mathbb{R}^{B\times C\times H\times W}$.

\subsubsection{Scale-Adaptive Strategy for Channel and Spatial Attentions}
The dual attention block merges the robust spatial feature extraction capabilities of the SCA module with the channel feature extraction strengths of the GCA module. The scale-adaptive strategy aims to dynamically adjust the attention focus based on the size of the bone features, enabling the two parallel branches to complement each other and allowing for a comprehensive interaction between channel and spatial dimensions. Practically, this operation can be implemented by ``nn.Parameter'' in PyTorch. This leverages the ability to treat a tensor as a trainable parameter and allows the model to learn how much to scale certain features during training. By learning the scale factor, the network can adapt its focus, making certain features more or less prominent depending on their contribution to accurate spine segmentation. We apply two learnable scale parameters to each parallel branch of the dual attention block and integrate the scale-adaptive mechanism with the scaling function, enabling SACSAM to capture the multi-scale characteristics of spinal images and emphasize features that are critical for spine segmentation regardless of their sizes. The computation of the scale-adaptive strategy is as allows:
\begin{equation}
\begin{aligned}
SACSAM(X) = \lambda_{1} X_{c} + \lambda_{2} X_{s}(X_{s})
\end{aligned}
\end{equation}
where $\lambda_{1}$ and $\lambda_{2}$ are learnable scale parameters that enable adaptive control of the importance of each branch for channel and spatial information in this spine segmentation task. $X_{s}(X_{s})$ means the feature processing through two consecutive SCA modules.

\subsection{Transformer Decoder with Cross-Attention} \label{decoder}
Multi-head self-attention has been proven to capture semantic-level affinity \cite{affinity} and effectively encode structural information of the spine. Inspired by DETR \cite{DETR}, each segment in an ultrasound image can be represented as a $d$-dimensional feature vector (``object query'') and processed by a Transformer decoder, trained with a set prediction objective. Therefore, we propose a Transformer decoder with cross-attention, which consists of six decoder blocks, to obtain long-range dependencies for better structure awareness. Figure \ref{SA2Net}(b) illustrates the architecture of this Transformer decoder. After feeding multi-scale image features from the backbone into SACSAM, the outputs are fully utilized to update query features in our proposed Transformer decoder. Firstly, to explicitly incorporate semantic information and enforce class-level reasoning, we design $N$ learnable class-specific queries $q_{c}$, where $N$ is the number of classes and each represents a target category. These queries serve as structural prototypes that enable the decoder to distinguish anatomically different bone features. Then, self-attention calculation is performed to enquire structural knowledge and reason long-range dependencies between local parts from the same bone region, obtaining more representative features. Next, we utilize the cross-attention mechanism to gather both semantic information of image features and class-wise structural information from $q_{c}$. Each query pulls semantically relevant information from SACSAM-enhanced image feature maps. This enforces class-aware selection of features and boosts the localization of class-specific regions in ultrasound VPI images. The process can be expressed as follows:
\begin{equation}
\begin{aligned}
f^{l} = Softmax (q_{c}^{l} {k^{l}}^{T}) v^{l} + f^{l-1}
\end{aligned}
\end{equation}
where $l$ is the layer index, $f^{l}$ refers to the output class-wise feature map at the $l^{th}$ layer. $k^{l}$ and $v^{l}$ are from two different linear transformations of the input image feature $f^{l-1}$, while $q_{c}^{l} \in \mathbb{R}^{N\times d_{h}}$ represents input $N$-class query features to the Transformer decoder, and $d_{h}$ is the hidden dimension.

By interacting with SACSAM-enhanced image features through multi-layer cross-attention, these queries guide the generation of semantic class masks with better structural integrity. The output feature from the final layer of the Transformer decoder is utilized for predicting semantic class masks $m_{n} \in \mathbb{R}^{d_{m}}, 1\leq n\leq N$, where $d_{m}$ is the class mask dimension, and $1\leq n\leq N$ denotes class $n$. The masks are calculated through a 3-layer multi-layer perceptron (MLP):
\begin{equation}
\begin{aligned}
m_{n} = MLP(f_{n} ^{-1})
\end{aligned}
\end{equation}

Furthermore, the obtained class-wise masks form the foundation for structure-affinity transformation (see Section \ref{sat}), where they drive the computation of class-specific affinity matrices and structure-aware attention maps.

\begin{figure}[t]
\centerline{\includegraphics[keepaspectratio, width=0.4 \columnwidth]{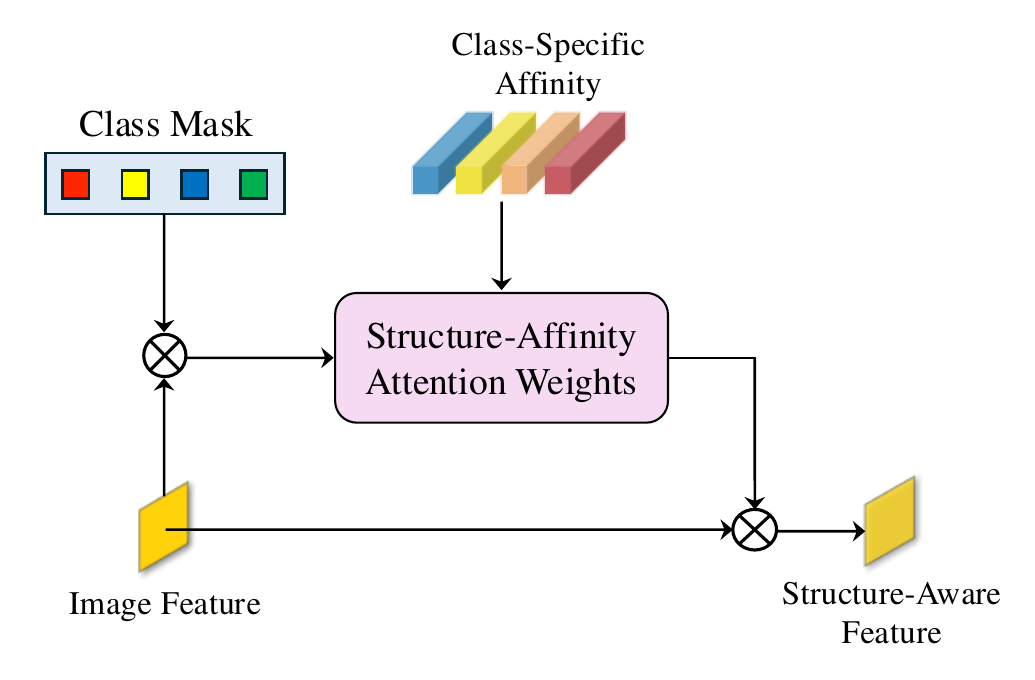}}
\caption{The overall structure of the Structure-Affinity Transformation module. The key is to calculate structure-affinity attention weights for structure-aware feature, with pixel-level classification confidence applied to class-specific affinity.}
\vspace{-5mm}
\label{SAM}
\end{figure}

\subsection{Structure-Affinity Transformation} \label{sat}
In a spinal image, three spine bones are typically identified: rib, thoracic process, and lump. These bone features exhibit a relatively consistent shape and position across different spinal images, providing valuable prior knowledge regarding their structural attributes and spatial relationships. We also notice the consistency between multi-head self-attention in Transformers and semantic-level affinity. To capture this rich information, as shown in Figure \ref{SAM}, we propose structure-affinity transformation to learn and encode structural information into attention maps, supervised by class-specific spine bone affinity. The core idea is to predict structure-affinity attention weights for each category by inferring the relationship between pixel-level classification confidence and affinity produced from the previous Transformer decoder. While attention-based affinity learning have been previously explored in semantic segmentation \cite{affinity}, our structure-affinity transformation introduces a novel mechanism specifically tailored to spinal ultrasound VPI image segmentation. Unlike prior work, we generate class-specific affinity matrices using output features of the Transformer decoder, and dynamically weight them based on classification confidence vectors. This class-specific modeling approach infers structure-affinity attention weights not just from local feature similarity, but from structural consistency derived from realistic bone anatomy. By incorporating classification confidence and re-weighting features through spine bone affinity-guided attention, our module enables explicit encoding of anatomical structure priors and enhances the separability of adjacent tissue parts and segment boundaries in complex ultrasound VPI images. In our implementation, considering the categories of bone features and background, we need four attention maps to encapsulate the structural knowledge and enhance the contextual information of bone features, i.e., $N=4$ (three bone features and the background).

Let there be $N$ classes as mentioned in Section \ref{decoder}. We utilize the output class-wise features from the multi-layer Transformer decoder to produce class-specific affinity $a_{n} ^{l}$ by employing an MLP layer: $a_{n} ^{l} = MLP(f_{n} ^{l})$. Essentially, the self-attention mechanism functions as a directed graphical model \cite{affinity}, where the affinity matrix aligns with the attention map, as points sharing the same structural knowledge are assumed to obtain equal affinity. Consequently, with the learned reliable spine bone affinity, the propagation process can diffuse the bone regions with high affinity and dampen the wrongly activated tissue parts, ensuring that the attention maps align better with segment boundaries.

Meanwhile, we perform a dot product operation on class mask $m_{n}$ and image feature $f^{l}$ to calculate the confidence vector $j_{n}^{l} = m_{n} \cdot f^{l}$, where $0\leq j_{n}^{l}\leq 1$, indicating the confidence of a pixel at layer $l$ of the Transformer decoder belonging to class $n$. Each mask $m_{n}$ represents the comprehensive feature of class $n$, and the dot product between $m_{n}$ and the image feature measures the similarity between the class and the pixels. After acquiring this classification confidence vector $j^{l} = [j_{1} ^{l},j_{2} ^{l},\ldots,j_{N} ^{l}]$, where $\sum _{n} j_{n}^{l} = 1$, we predict structure-affinity attention weights for each class at layer $l$ by applying the confidence vector to class-specific affinity $a_{n} ^{l}$. Next, a matrix multiplication between the attentive affinity matrix and the image feature results in the structure-aware feature $\hat{f^{l}}$. The transformation is summarized as follows:
\begin{equation}
\begin{aligned}
j^{l} = [j_{1} ^{l},j_{2} ^{l},\ldots,j_{N} ^{l}], j_{n}^{l} = m_{n} \cdot f^{l}
\end{aligned}
\end{equation}
\begin{equation}
\begin{aligned}
A^{l} = \sum _{n} j_{n}^{l} a_{n}^{l}
\end{aligned}
\end{equation}
\begin{equation}
\begin{aligned}
\hat{f^{l}} = A^{l} f^{l}
\end{aligned}
\end{equation}

This operation transforms semantic features with the same category towards similar bone structures, thus pulling them closer together. On the contrary, for features belonging to different categories, their discrepancy in spine bone structures pushes them further apart. This comprehensive acquisition of structural knowledge of different spine bones is facilitated by structure-affinity attention weights because the output features are directly synthesized with the class-specific affinity.

\subsection{Structure-Aware Module (SAM)}
The design of the structure-aware module is to integrate the structure-affinity transformation module into a multi-layer Transformer module that can be implemented in most encoder-decoder segmentation architecture. We adopt this module to jointly infer semantic class masks and structure-affinity attention weights with learnable class queries. Each class query represents a category and inquires class-specific structural information in semantic feature representations. Then, the generated semantic class masks are used to perform more accurate and flexible pixel-level classification, while structure-affinity attention weights are applied to transform image features.

\subsection{Feature Mixing Loss and Outputs Aggregation}
After processing the features extracted from the backbone using our proposed SACSAM and SAM modules, we feed cross-dimensional features and class-specific structure-affinity features into a classifier to produce multiple segmentation maps. These maps are then aggregated to generate the final prediction map $\hat{y}$ for multi-class segmentation. In this scheme, we adopt a combinatorial learning strategy to enable better model training. This involves taking all the segmentation maps from different parts of SA$^{2}$Net as inputs and calculating the loss for all possible combinations of predictions, including $2^{k} - 1$ non-empty subsets of $k$ prediction maps, and then summing these losses. Considering that there are two prediction maps generated from SACSAM and SAM, respectively, i.e., $k=2$, the feature mixing loss aggregation approach produces a total of $2^{2} - 1 = 3$ maps. By utilizing these three prediction maps, including the two original maps and a mixed map based on combining the two original maps, and mixing features from the decoder during loss calculation, this strategy creates new synthetic predictions and improves the performance of spine segmentation. The optimized loss $\mathcal{L} _{total}$ can be expressed as follows:
\begin{equation}
\begin{aligned}
\mathcal{L} _{total} = \alpha \mathcal{L} _{p1+p2} + \beta \mathcal{L} _{p1} + \gamma \mathcal{L} _{p2}
\end{aligned}
\end{equation}
where $\mathcal{L} _{p1}$ and $\mathcal{L} _{p2}$ are the losses of each individual segmentation map from the SAM and SACSAM modules, respectively. $\mathcal{L} _{p1+p2}$ denotes the segmentation loss on the predicted pixel-wise label $\hat{y}$. In order to enhance the classification ability for each pixel, we choose the Cross Entropy (CE) loss function to calculate the classification error of each pixel. $\alpha$, $\beta$, and $\gamma$ are the weights assigned to each loss to control the trade-off between the loss terms. Empirically, we set $\alpha=1.0$, $\beta=0.4$, and $\gamma=0.5$ in this paper.

\section{Experiments} \label{experiment}
In this section, we conduct extensive experiments to demonstrate the superiority of our proposed SA$^{2}$Net architecture. We introduce the experimental settings, including the preparation of the collected dataset, evaluation metrics, and the implementation of the proposed architecture, followed by a comparative analysis of SA$^{2}$Net against other state-of-the-art (SOTA) models.

\subsection{Dataset}
The spinal ultrasound VPI data are obtained from 3D ultrasound scanning of the whole spine region using the Scolioscan system (Model SCN801, Telefield Medical Imaging Ltd., Hong Kong). A total of 109 patients are selected, with their data collected from authentic clinical practices. These patients, with an average age of 15.6 $\pm$ 2.7 years, suffer from varying degrees of spinal deformity during their adolescent growth period. In terms of gender distribution, there are 82 females and 27 males. For model development, three experts, two with 5 years of ultrasound experiments and one with more than 2 years of experience, manually annotate the bone features to serve as the ground-truth segments. Therefore, this dataset is representative and faithfully mirrors the spinal profiles of patients with spinal deformity encountered in real clinical practice.

In this dataset, a total of 109 cases are included, with the 3-fold subject-independent cross-validation employed. The whole dataset is randomly split into 3 folds. Cases from a single fold are retained to evaluate the performance of the model and the other 2 folds are used for training. The cross-validation procedure is repeated 3 times, once for each fold, and the results are averaged over the 3 rounds to obtain the final estimation. Furthermore, these 2D VPI images have a resolution of about 2600 $\times$ 640 pixels. To ensure uniformity, all images are resized to 2048 $\times$ 512 pixels. During the training stage, we crop the image of a size of 512 $\times$ 512 pixels as the input for SA$^{2}$Net. In the testing stage, the samples, which retain the resized resolution of 2048 $\times$ 512 pixels, are passed into SA$^{2}$Net to generate the segmentation maps.

\subsection{Evaluation Metrics}
In evaluating the performance of SA$^{2}$Net, we utilize a comprehensive set of metrics, including Dice Similarity Score (DSC), Intersection over Union (IoU), and Pixel Accuracy (Acc). These metrics provide a deep understanding of the network's accuracy, precision, and robustness, which are calculated as follows:
\begin{equation}
\begin{aligned}
DSC(Y,\hat{Y}) = \frac{2 \times \left| Y \cap \hat{Y} \right|}{\left| Y \right| + \left| \hat{Y} \right|} \times 100\%
\end{aligned}
\end{equation}
\begin{equation}
\begin{aligned}
IoU(Y,\hat{Y}) = \frac{\left| Y \cap \hat{Y} \right|}{\left| Y \cup \hat{Y} \right|} \times 100\%
\end{aligned}
\end{equation}
\begin{equation}
\begin{aligned}
Acc = \frac{TP + TN}{TP + TN + FP + FN} \times 100\%
\end{aligned}
\end{equation}
where $Y$ and $\hat{Y}$ are the ground-truth mask and predicted segmentation maps, respectively, while TP, TN, FP, and FN refer to true positive, true negative, false positive, and false negative points, respectively.

\subsection{Implementation Details}
\subsubsection{Network Structure}
As depicted in Figure \ref{SA2Net}(a), the choice of backbone is flexible and SA$^{2}$Net is compatible with any backbone architecture. In our study, we employ the highly-regarded baseline for medical image segmentation, UNet \cite{UNet}, the standard convolution-based ResNet \cite{Zixun_joint46} backbones (R50 with 50 layers), and the recently proposed Transformer-based Swin-Transformer \cite{swin} backbone to showcase the superior spine segmentation performance and great generalization ability of our proposed model.

\subsubsection{Training Settings}
We use PyTorch 1.13.1 with CUDA 11.7 in all of our experiments. Our implementation is based on MMSegmentation libraries \cite{mmseg2020}. All models are trained on a single NVIDIA RTX 4090 GPU with 24GB of memory. To enhance the robustness of the model, we perform three data augmentation techniques: random scale jittering from the range (0.5$\sim$2.0), random cropping, and random flipping. For the input VPI data, we use a crop size of 512 $\times$ 512, a batch size of 4, and train all models for 160k iterations. All models are trained using the AdamW \cite{maskformer_31} optimizer and the poly \cite{maskformer_7} learning rate schedule with an initial learning rate of 10$^{-4}$, a momentum of 0.9, and a weight decay of 10$^{-4}$ for regularization. We report the performance of multi-scale inference with flip and scales of 0.5, 0.75, 1.0, 1.25, 1.5, 1.75.

\subsection{Comparison with the State-of-the-Arts}
We conduct a comprehensive comparison between our proposed SA$^{2}$Net and other state-of-the-art models on spinal ultrasound VPI images under the same setting and experimental environment. These compared methods are mainly divided into two categories: convolutional networks and Transformer-based networks. For CNN backbones, we select benchmark methods such as FCN \cite{FCN}, PSPNet \cite{PSPNet}, DeepLabv3 \cite{DeepLab}, and nnUNet \cite{nnUNet}, which are based on U-shaped architectures for medical image segmentation, as well as previously explored ResNet-based methods like SEAM \cite{Zixun_joint24} and SADANet \cite{BIBM}, especially designed for spine segmentation from ultrasound VPI images. Meanwhile, recent Vision Transformer \cite{ViT} backbones (e.g., SATR \cite{ISBI} and Swin-Transformer \cite{swin}) are adopted to report the spine segmentation performance.

\begin{table*}[t]
\caption{Performance comparison of the proposed method against the SOTA approaches on the spine segmentation task. We report inference results with $\pm$Standard Deviation (Std) and efficiency comparison. ``FPS'' is measured as the number of VPI images processed per second during inference. The best results are highlighted in bold and the second-best results are underlined.}
\Huge
\renewcommand\arraystretch{2.0}
\centering
\resizebox{0.96 \linewidth}{!}{
\begin{tabular}{c|c|c|ccc|ccc|ccc|ccc|ccc}
\hline
\multirow{2}{*}{}                 & \multirow{2}{*}{Method}       & \multirow{2}{*}{Backbone} & \multicolumn{3}{c|}{Rib}                                        & \multicolumn{3}{c|}{Thoracic}                                   & \multicolumn{3}{c|}{Lump}                                       & \multicolumn{3}{c|}{Ave.}                                        & \multicolumn{3}{c}{Efficiency} \\ \cline{4-18} 
&               &                           & DSC                 & IoU                 & Acc                 & DSC                 & IoU                 & Acc                 & DSC                 & IoU                 & Acc                 & DSC                 & IoU                 & Acc                 & Train & FPS & GPU \\ \hline
\multirow{7}{*}{\rotatebox{90}{CNN Backbones}}
& DeepLabv3 \cite{DeepLab}      & \multirow{3}{*}{UNet}     & 78.40$\pm$0.05               & 64.48$\pm$0.30               & 76.64$\pm$0.26               & 76.43$\pm$0.22               & 61.85$\pm$0.28               & 70.94$\pm$0.38               & 84.24$\pm$0.15               & 72.77$\pm$0.27               & 81.02$\pm$0.27               & 83.83$\pm$0.04               & 72.96$\pm$0.50               & 81.47$\pm$0.22               & 19.7h & 1.5 & 13.1G \\
& PSPNet \cite{PSPNet}          &                           & 78.81$\pm$0.06               & 65.03$\pm$0.46               & 76.38$\pm$0.11               & 77.09$\pm$0.14               & 62.73$\pm$0.15               & 73.09$\pm$0.16               & 83.56$\pm$0.31               & 71.77$\pm$0.05               & 78.96$\pm$0.07               & 84.08$\pm$0.33               & 73.26$\pm$0.18               & 81.84$\pm$0.07               & 28.0h & 1.4 & \textbf{12.2G} \\
& FCN \cite{FCN}                &                           & \underline{79.53$\pm$0.22}   & \underline{66.02$\pm$0.31}   & 78.11$\pm$0.25               & \underline{77.76$\pm$0.03}   & 63.61$\pm$0.04               & 74.04$\pm$0.06               & 84.67$\pm$0.10               & 73.27$\pm$0.14               & 82.0$\pm$0.08                & \underline{84.59$\pm$0.05}   & 74.02$\pm$0.07               & 82.88$\pm$0.21               & \textbf{14.0h} & 1.7 & \underline{12.8G} \\ \cline{2-3}
& nnUNet \cite{nnUNet}          & --                        & 78.45$\pm$0.11               & 65.73$\pm$0.47               & \textbf{80.0$\pm$0.02}       & 77.30$\pm$0.36               & 63.07$\pm$0.07               & 77.81$\pm$0.09               & 83.64$\pm$0.20               & 74.72$\pm$0.35               & 81.59$\pm$0.04               & 84.38$\pm$0.35               & 71.48$\pm$0.46               & 82.28$\pm$0.16               & \underline{15.6h} & 0.3 & 16.0G \\ \cline{2-18} 
& SEAM \cite{Zixun_joint24}     & \multirow{2}{*}{R50}      & 77.92$\pm$0.27               & 65.64$\pm$0.25               & 78.80$\pm$0.08               & 76.60$\pm$0.36               & \underline{63.92$\pm$0.27}   & 72.25$\pm$0.13               & 84.40$\pm$0.04               & \underline{75.72$\pm$0.35}   & 83.48$\pm$0.05               & 83.92$\pm$0.37               & 72.88$\pm$0.02               & 82.29$\pm$0.07               & 16.2h & \underline{1.8} & 15.3G \\
& SADANet \cite{BIBM}           &                           & 78.49$\pm$0.05               & 64.61$\pm$0.25               & 79.18$\pm$0.27               & \underline{77.76$\pm$0.13}   & 63.61$\pm$0.07               & \textbf{78.80$\pm$0.08}      & \textbf{86.29$\pm$0.06}      & \textbf{75.92$\pm$0.05}      & \textbf{85.09$\pm$0.09}      & 84.50$\pm$0.05               & \underline{74.29$\pm$0.07}   & \textbf{84.56$\pm$0.14}      & 18.7h & \textbf{1.9} & 19.7G \\ \cline{2-18}
& SA$^{2}$Net (Ours)            & R50                       & \textbf{80.43$\pm$0.13}      & \textbf{67.08$\pm$0.19}      & \underline{79.24$\pm$0.50}   & \textbf{78.52$\pm$0.07}      & \textbf{64.63$\pm$0.10}      & \underline{78.19$\pm$0.56}   & \underline{85.88$\pm$0.12}   & 75.26$\pm$0.18               & \underline{83.69$\pm$0.32}   & \textbf{85.21$\pm$0.01}      & \textbf{74.92$\pm$0.01}      & \underline{84.19$\pm$0.42}   & 20.0h & \textbf{1.9} & 18.6G \\ \hline
\multirow{6}{*}{\rotatebox{90}{Transformer Backbones}}
& SETR \cite{SETR}                    & ViT                       & 80.34$\pm$0.34               & 67.14$\pm$0.47               & 80.29$\pm$0.10               & 78.46$\pm$0.11               & 64.56$\pm$0.31               & 77.49$\pm$0.04               & 86.48$\pm$0.13               & 76.18$\pm$0.30               & 86.99$\pm$0.40               & 85.44$\pm$0.48               & 75.27$\pm$0.05               & 85.32$\pm$0.10               & \underline{20.0h} & 0.5 & 20.0G \\ \cline{2-3}
& SATR \cite{ISBI}                    & --                        & 80.92$\pm$0.19               & 67.95$\pm$0.38               & 80.84$\pm$0.30               & 79.14$\pm$0.03               & 65.31$\pm$0.29               & 77.99$\pm$0.08               & 87.03$\pm$0.01               & 77.04$\pm$0.24               & 89.54$\pm$0.04               & 85.81$\pm$0.18               & 75.81$\pm$0.30               & \underline{86.59$\pm$0.03}   & 23.7h & 0.8 & 19.4G \\ \cline{2-18} 
& \multirow{2}{*}{UPerNet \cite{u}}   & Swin-B                    & 80.86$\pm$0.23               & 67.88$\pm$0.31               & 80.06$\pm$0.35               & 78.90$\pm$0.35               & 65.16$\pm$0.47               & 77.56$\pm$0.0.18             & 87.82$\pm$0.15               & 78.28$\pm$0.24               & 89.13$\pm$0.33               & 86.17$\pm$0.04               & 76.39$\pm$0.04               & 85.99$\pm$0.28               & \textbf{16.7h} & 0.8 & \textbf{14.1G} \\
&                                     & Swin-L                    & 81.17$\pm$0.18               & 68.31$\pm$0.25               & 79.46$\pm$0.02               & \underline{79.61$\pm$0.03}   & \underline{66.13$\pm$0.04}   & 78.65$\pm$0.19               & \underline{87.93$\pm$0.32}   & 78.47$\pm$0.40               & 87.69$\pm$0.20               & 86.39$\pm$0.12               & \underline{76.68$\pm$0.19}   & 86.02$\pm$0.36               & 21.7h & 0.6 & 19.0G \\ \cline{2-18} 
& \multirow{2}{*}{SA$^{2}$Net (Ours)} & Swin-B                    & \underline{81.52$\pm$0.04}   & \underline{68.81$\pm$0.06}   & \underline{81.59$\pm$0.13}   & 79.39$\pm$0.08               & 65.82$\pm$0.12               & \underline{78.90$\pm$0.52}  & 87.88$\pm$0.10               & \underline{78.54$\pm$0.16}   & \underline{89.78$\pm$0.13}   & \underline{86.41$\pm$0.03}   & 76.67$\pm$0.04               & 86.31$\pm$0.21               & 21.0h & \textbf{1.3} & \underline{15.8G} \\
&                                     & Swin-L                    & \textbf{81.80$\pm$0.24} & \textbf{69.21$\pm$0.35} & \textbf{81.57$\pm$0.09} & \textbf{79.88$\pm$0.01} & \textbf{66.51$\pm$0.01} & \textbf{79.75$\pm$0.06} & \textbf{88.47$\pm$0.07} & \textbf{79.33$\pm$0.10} & \textbf{90.21$\pm$0.35} & \textbf{86.71$\pm$0.10} & \textbf{77.18$\pm$0.15} & \textbf{86.84$\pm$0.01} & 23.0h & \underline{1.0} & 20.9G \\ \hline
\end{tabular}}
\vspace{-2mm}
\label{quantity}
\end{table*}

\begin{figure*}[t]
\centering
\includegraphics[width=\linewidth]{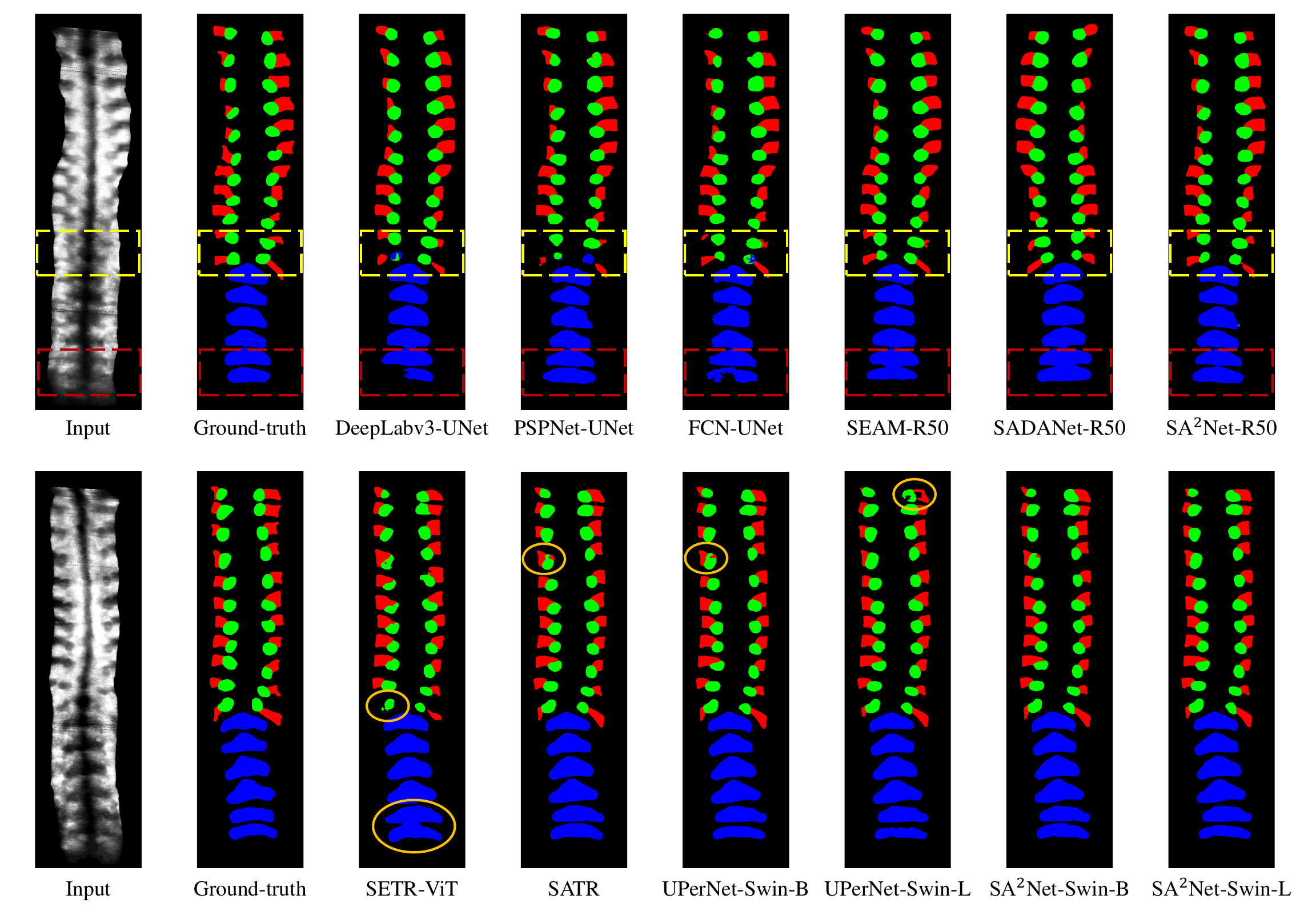}
\caption{Qualitative spine bone segmentation comparisons on ultrasound VPI images based on different methods. The segmented rib, thoracic process, and lump are annotated in red, green, and blue, respectively. The yellow rectangular box highlights the area around the boundary of the thoracic and lumbar region, while the red rectangular box highlights a part of the lumbar vertebra. The orange circle marks the defect parts of the segmentation results.}
\vspace{-5mm}
\label{visual}
\end{figure*}

\subsubsection{Quantitative Comparison}
In the experiment, we conduct an overall evaluation of different comparison methods in terms of average and individual segmentation performance of three spine bone features: rib, thoracic process, and lump, along with computational efficiency. Table \ref{quantity} presents the quantitative results of the proposed SA$^{2}$Net and current mainstream convolutional and Transformer-based networks. With the same standard CNN backbones (e.g., R50), our SA$^{2}$Net achieves the highest average DSC (85.21\%) and IoU (74.92\%) metrics, surpassing the existing best U-shaped architecture, with an improvement of nearly 1\% on all the evaluation metrics. Compared with ResNet-based methods, although SA$^{2}$Net outperforms SEAM \cite{Zixun_joint24} by a notable margin, it only secures the second-best average Acc metric (84.19\%), which is slightly lower than our previously proposed method, SADANet \cite{BIBM}. When considering the evaluation metrics of individual segments, it is clear that the proposed SA$^{2}$Net significantly outperforms all CNN-based SOTA methods in two out of three bone features, i.e., rib and thoracic process. This demonstrates the effectiveness of the proposed scale-adaptive structure-affinity transformation for spine bone segmentation.

More importantly, SA$^{2}$Net is also compatible with Transformer backbones, producing a new state-of-the-art performance with two Swin-Transformer variants (Swin-B and Swin-L). Swin-B has four stages of hidden layers, each with 2, 2, 18, 2 layer numbers, respectively. And Swin-L is the version of about $2\times$ the model size, compared with Swin-B \cite{swin}. Surprisingly, among all Transformer-based methods, SA$^{2}$Net-Swin-L achieves the best average DSC of 86.71\%, IoU of 77.18\% and Acc of 86.84\%, beating the nearest SOTA method UPerNet-Swin-L by 0.32\%, 0.50\%, and 0.82\%, respectively. Meanwhile, SA$^{2}$Net-Swin-B exhibits excellent performance in spine segmentation, securing second place. It is worth noting that not only the mean metrics but also the category metrics consistently show significant improvement. We attribute the superior segmentation capability of SA$^{2}$Net to the design of the scale-adaptive channel-spatial attention module and structure-aware module, as well as the introduction of feature mixing loss aggregation. 

In addition, the efficiency comparison in Table \ref{quantity} also displays higher inference speed of our SA$^{2}$Net with both CNN and Transformer backbones, which is competitive with other state-of-the-art methods. Specifically, SA$^{2}$Net-R50 achieves the highest inference efficiency of 1.9 FPS, validating the possibility of SA$^{2}$Net for real-world applications in spine segmentation from ultrasound VPI images.

\subsubsection{Qualitative Comparison}
To further illustrate the better segmentation performance of our proposed method, we visualize two samples from the testing part with different segmentation models based on CNN and Transformer backbones. Figure \ref{visual} presents qualitative results on spinal ultrasound VPI images. For convolutional networks, it can be observed from the red rectangular box that nearly all methods face challenges in segmenting the lumbar vertebra. However, our SA$^{2}$Net performs exceptionally well in segmenting the lumbar vertebra. Another challenging area is around the boundary of the thoracic and lumbar region, as shown in the yellow rectangular box. The U-shaped architecture struggles to identify spine bone features in this challenging area. Furthermore, although ResNet-based methods, i.e., SEAM \cite{Zixun_joint24} and SADANet \cite{BIBM}, manage to distinguish the rib and thoracic process, our proposed SA$^{2}$Net provides a more accurate and smoother shape of each spine bone feature, closely resembling the ground-truth segmentation masks. Similarly, qualitative comparisons from the second row of Figure \ref{visual} demonstrate that existing Transformer-based SOTA methods obfuscate the boundary demarcation lines and yield unsatisfactory results at the connection area between the rib and thoracic process (see orange circles). In contrast, SA$^{2}$Net is able to predict clearer and more appealing segmentation masks in this region. From these qualitative results, the superiority of our proposed SA$^{2}$Net is evident.

\begin{table}[t]
\caption{Effect of different components of SA$^{2}$Net with Swin-L backbone on the spinal ultrasound VPI data. ``Params'' refers to the number of parameters, while ``Flops'' is calculated under the input resolution of 512 $\times$ 512. The best results are highlighted in bold and the second-best results are underlined.}
\Huge
\renewcommand\arraystretch{1.3}
\centering
\resizebox{0.5 \linewidth}{!}{
\begin{tabular}{ccc|cc|ccc}
\hline
\multicolumn{3}{c|}{Components}                                           & \multirow{2}{*}{Params (M)} & \multirow{2}{*}{Flops (G)} & \multicolumn{3}{c}{Ave.}                                       \\ \cline{1-3} \cline{6-8} 
\multicolumn{1}{c|}{SACSAM} & \multicolumn{1}{c|}{SAM} & Loss Aggregation &                             &                            & DSC                & IoU                 & Acc                 \\ \hline
\multicolumn{1}{c|}{\XSolidBrush}     & \multicolumn{1}{c|}{\XSolidBrush}  & \XSolidBrush     & 0                           & 0                          & 86.39$\pm$0.12         & 76.68$\pm$0.19          & 86.02$\pm$0.36          \\
\multicolumn{1}{c|}{\Checkmark}       & \multicolumn{1}{c|}{\XSolidBrush}  & \XSolidBrush     & 5.45                        & 94.63                      & 86.41$\pm$0.07         & 76.64$\pm$0.11          & 86.07$\pm$0.03          \\
\multicolumn{1}{c|}{\XSolidBrush}     & \multicolumn{1}{c|}{\Checkmark}    & \XSolidBrush     & 6.56                        & 86.3                       & 86.41$\pm$0.17         & 76.69$\pm$0.23          & 86.24$\pm$0.36          \\
\multicolumn{1}{c|}{\Checkmark}       & \multicolumn{1}{c|}{\Checkmark}    & \XSolidBrush     & --                          & --                         & \underline{86.61$\pm$0.01} & \underline{76.76$\pm$0.24} & \underline{86.45$\pm$0.11} \\
\multicolumn{1}{c|}{\Checkmark}       & \multicolumn{1}{c|}{\Checkmark}    & \Checkmark       & 12                          & 180.93                     & \textbf{86.71$\pm$0.10} & \textbf{77.18$\pm$0.15} & \textbf{86.84$\pm$0.01} \\ \hline
\end{tabular}}
\label{abla1}
\end{table}

\begin{table}[t]
\caption{Quantitative segmentation performance comparison of the scale-adaptive channel-spatial attention module (SACSAM) and its scale-adaptive strategy in SA$^{2}$Net with R50 backbone in terms of three spine bone regions. The best results are highlighted in bold and the second-best results are underlined.}
\Huge
\renewcommand\arraystretch{1.6}
\centering
\resizebox{0.84 \linewidth}{!}{
\begin{tabular}{c|ccc|ccc|ccc|ccc}
\hline
\multirow{2}{*}{Method} & \multicolumn{3}{c|}{Rib}                         & \multicolumn{3}{c|}{Thoracic}                    & \multicolumn{3}{c|}{Lump}                        & \multicolumn{3}{c}{Ave.}                         \\ \cline{2-13} 
                        & DSC            & IoU            & Acc            & DSC            & IoU            & Acc            & DSC            & IoU            & Acc            & DSC            & IoU            & Acc            \\ \hline
SCSA-Net \cite{SCSA-Net}& 78.43$\pm$0.13          & 64.51$\pm$0.19          & 77.27$\pm$0.05             & 77.22$\pm$0.07          & 62.88$\pm$0.10          & 76.07$\pm$0.12          & 85.37$\pm$0.12          & 74.47$\pm$0.18          & 82.38$\pm$0.31             & 84.38$\pm$0.01          & 73.75$\pm$0.01          & 84.05$\pm$0.16          \\

SADANet \cite{BIBM}     & 78.49$\pm$0.05          & 64.61$\pm$0.25          & \underline{79.18$\pm$0.27} & 77.76$\pm$0.13          & 63.61$\pm$0.07          & \textbf{78.80$\pm$0.08} & \textbf{86.29$\pm$0.06} & \textbf{75.92$\pm$0.05} & \underline{85.09$\pm$0.09} & 84.50$\pm$0.05          & 74.29$\pm$0.07          & \textbf{84.56$\pm$0.14} \\

$\sim$ w/o SA           & 78.92$\pm$0.06          & 65.31$\pm$0.19          & 78.42$\pm$0.27             & 77.99$\pm$0.04          & 63.98$\pm$0.05          & 76.80$\pm$0.11          & 85.35$\pm$0.38          & 74.79$\pm$0.26          & 84.91$\pm$0.07             & 84.63$\pm$0.06          & 74.30$\pm$0.33          & 84.11$\pm$0.27          \\

$\sim$ w/ SACSAM        & \underline{79.98$\pm$0.09} & \underline{66.64$\pm$0.01} & 78.37$\pm$0.04          & \underline{78.22$\pm$0.12} & \underline{64.23$\pm$0.16} & 75.48$\pm$0.01          & 85.87$\pm$0.10          & 75.23$\pm$0.16          & \textbf{85.20$\pm$0.06} & \underline{84.95$\pm$0.08} & \underline{74.53$\pm$0.11} & 84.01$\pm$0.05          \\
SA$^{2}$Net (Ours)      & \textbf{80.43$\pm$0.13} & \textbf{67.08$\pm$0.19} & \textbf{79.24$\pm$0.50} & \textbf{78.52$\pm$0.07} & \textbf{64.63$\pm$0.10} & \underline{78.19$\pm$0.56} & \underline{85.88$\pm$0.12} & \underline{75.26$\pm$0.18} & 83.69$\pm$0.32          & \textbf{85.21$\pm$0.01} & \textbf{74.92$\pm$0.01} & \underline{84.19$\pm$0.42} \\ \hline
\end{tabular}}
\vspace{-2mm}
\label{abla2}
\end{table}

\subsection{Ablation Study}
In order to fully validate the effectiveness of each component in our proposed SA$^{2}$Net, we conduct a set of ablation experiments on the spinal ultrasound VPI images, detailed in Table \ref{abla1}. We assess the model performance by removing or adding SACSAM, SAM, and loss aggregation to understand their effects. It is worth noting that without the introduction of feature mixing loss aggregation, the model is trained with $\alpha=0$. When only adding the SACSAM or SAM component, the model directly adopts a single learning strategy to process the features extracted from the Swin-Transformer \cite{swin} backbone, i.e., $\beta=1.0$ or $\gamma=1.0$. Table \ref{abla1} reveals the results of our SA$^{2}$Net with or without SACSAM, SAM, and loss aggregation. It can be seen that both SACSAM and SAM show great segmentation performance, with the incorporation of SAM proving to be more effective, achieving better mean IoU (76.64-76.69\%) and Acc (86.07-86.24\%). When combining these two modules, it produces the second-highest average DSC of 86.61\%, IoU of 76.76\%, and Acc of 86.45\%. Furthermore, the feature mixing loss aggregation approach, as a special augmentation method, significantly enhances segmentation performance, achieving the best mean evaluation metrics. It is evident that each component contributes to the overall network.

Besides ablations on components of SA$^{2}$Net, we further explore the superiority of the scale-adaptive channel-spatial attention module by only adding the SACSAM component in our proposed SA$^{2}$Net, denoted as ``$\sim$ w/ SACSAM''. We also remove learnable scale parameters $\lambda_{1}$ and $\lambda_{2}$ in the SACSAM component and only retain parallel branches for channel and spatial information to investigate the effectiveness of the scale-adaptive strategy, labeled as ``$\sim$ w/o SA''. As tabulated in Table \ref{abla2}, we choose the conventional dual attention network of SCSA-Net \cite{SCSA-Net} and our previous method of SADANet \cite{BIBM}, then report the quantitative segmentation results with the ResNet backbone in terms of three spine bone features. Compared with other dual attention-based methods, a significant improvement can be observed not only on the average evaluation metrics but also on the metrics of individual segments, e.g., rib and thoracic process, when adding the SACSAM component to capture the multi-scale characteristics of spinal images. Surprisingly, the incorporation of SACSAM produces the best Acc of 85.20\% in the lumbar region, which reveals the powerful ability of the scale-adaptive channel-spatial attention module to achieve cross-dimensional global modeling of spinal images in the spine segmentation task. 
Meanwhile, the absence of the scale-adaptive strategy in SACSAM leads to a consistent decrease across all DSC and IoU evaluation metrics, especially in the rib region with a drop of more than 1\% on these two metrics. This shows that the scale-adaptive mechanism enhances the ability of the network to adaptively focus on contextually relevant features between the channel and spatial dimensions.

\section{Discussions and Limitations} \label{discussion}
We have demonstrated the state-of-the-art segmentation performance of our SA$^{2}$Net. One of the most critical clinical value of the proposed method is its ability to provide more robust and precise segmentation of the spine from ultrasound VPI images, ensuring that each bone feature is segmented accurately. By improving the segmentation accuracy, SA$^{2}$Net can aid in identifying extremely small deviations in the spine and detecting subtle spinal deformities that may otherwise be missed. This process is crucial for the proper assessment of spine curvature, particularly in calculating the spinal curve angle, allowing clinicians to make data-driven decisions based on highly accurate and detailed information of the spine. This can lead to more tailored and appropriate treatment plans, ensuring that patients receive the best care for their specific condition.

Additionally, traditional methods of spine segmentation rely on manual efforts, which are time-consuming, especially when dealing with large patient volumes. With our proposed SA$^{2}$Net, the entire spine segmentation and curvature assessment process is automated, enabling intelligent analysis of ultrasound VPI images. This significantly reduces the time and manual effort required by clinicians to diagnose scoliosis, which is especially beneficial in real-word clinical settings with high patient volumes. In busy clinical environments, where quick decision-making and efficient workflows are critical, this automation can speed up treatment planning, enabling doctors to provide faster interventions. This has great potential to impact public health, particularly in regions where large-scale monitoring is critical for early diagnosis.

However, the proposed model requires considerable computational resources during training, especially when it is implemented with high-capacity Transformer backbones, like Swin-L \cite{swin}. As shown in Table \ref{quantity}, SA$^{2}$Net-Swin-L takes approximately 23.0 hours to train and consumes up to 20.9 GB of GPU memory. Although its inference speed remains acceptable, such resource demands may pose challenges for the deployment in low-resource clinical environments, indicating potential for further improvements on lightweight network design. Besides, the current implementation is trained on a relatively limited spinal ultrasound VPI dataset with 109 cases. Although data augmentation and sufficient cross-validation approaches are used to mitigate overfitting, the constrained data size may still impact the scalability of SA$^{2}$Net to more diverse anatomical variations or imaging conditions. One potential solution is incorporating semi-supervised or self-supervised learning strategies in future work to enhance the accessibility of SA$^{2}$Net in broader clinical workflows.

\section{Conclusion} \label{conclusion}
In this paper, we propose a novel architecture, SA$^{2}$Net, that combines a scale-adaptive dual channel-spatial attention module (SACSAM) and a Transformer decoder-based structure-aware module (SAM) with structure-affinity transformation as the core, guided by a feature mixing loss and outputs aggregation approach for effective spine segmentation. The introduced scale-adaptive complementary strategy enables the dual attention block to fully capture the cross-dimensional correlation and adaptively learn important spine contextual information between the channel and spatial dimensions. The built structure-aware module infers semantic class masks and imposes structure-affinity attention weights on the segmented bone features with learnable queries. Compared with mainstream CNN and Transformer-based networks, the experiments reveal the superiority of our SA$^{2}$Net, significantly surpassing other state-of-the-art methods. We also prove the generalization ability of SA$^{2}$Net, showing that it can be easily integrated into any encoder-decoder segmentation architecture. Thanks to the superior segmentation performance and the great adaptability to various backbones, we believe that our proposed SA$^{2}$Net will be a valuable asset in clinical scoliosis diagnosis in the future.





\printcredits

\section*{Declaration of competing interest}
The authors declare that they have no known competing financial interests or personal relationships that could have appeared to influence the work reported in this paper.

\section*{Acknowledgments}
This work was supported by a grant from the Research Grants Council of the Hong Kong Special Administrative Region, China, under the Project Account Code B-Q86J, and the Research Institute for Smart Ageing, The Hong Kong Polytechnic University.

\section*{Data availability}
This study involves human subjects, and the data have been used are collected from authentic clinical practices. To ensure patient privacy, the relevant data are confidential and copyrighted.

\bibliographystyle{cas}

\bibliography{cas}



\end{document}